\documentclass[twoside]{article}
\pdfoutput=1

\usepackage{epsfig}
\usepackage{times}
\usepackage{latexsym}

\usepackage{subfigure}

\usepackage{diagbox} 
\usepackage{makecell}
\usepackage{enumitem}
\usepackage[english]{babel} 
\usepackage{rotating}
\usepackage{textpos}

\usepackage[utf8]{inputenc} 
\usepackage[numbers]{natbib}
\usepackage[T1]{fontenc}    
\usepackage{hyperref}       
\usepackage{url}            
\usepackage{booktabs}       
\usepackage{amsfonts}       
\usepackage{nicefrac}       
\usepackage{microtype}      
\usepackage{xcolor}         

\usepackage{graphicx}
\usepackage{mathrsfs}
\usepackage{amsmath,amssymb}
\usepackage[ruled,linesnumbered,noend]{algorithm2e}
\usepackage{multirow}
\usepackage{colortbl}

\usepackage{amsbsy,textcomp,marvosym,caption,threeparttable,amsthm,subfigure,float,lastpage,lscape}

\usepackage{eurosym,mathrsfs,fancyhdr,CJK,multicol,graphics,indentfirst,color,bm,upgreek,booktabs,graphicx,multirow,warpcol}
\def\footnoterule{\kern 1mm \hrule width 10cm \kern 2mm}

\allowdisplaybreaks
\catcode`@=11
\def\title#1{\vspace{3mm}\begin{flushleft}\vglue-.1cm\Large\bf\boldmath\protect\baselineskip=18pt plus.2pt minus.1pt #1
\end{flushleft}\vspace{1mm} }
\def\author#1{\begin{flushleft}\normalsize #1\end{flushleft}\vspace*{-4pt} \vspace{3mm}}

\def\jz#1#2{{$^{\footnotesize\textcircled{\tiny #1}}$\let\thefootnote\relax\footnotetext{\!\!$^{\footnotesize\textcircled{\tiny #1}}$#2}}}
\catcode`@=11
\def\section{\@startsection{section}{1}{\z@}%
 {-3ex \@plus -.3ex \@minus -.2ex}%
 {2.2ex \@plus.2ex}%
{\normalfont\normalsize\protect\baselineskip=14.5pt plus.2pt minus.2pt\bfseries}}
\def\subsection{\@startsection{subsection}{2}{\z@}%
 {-3ex\@plus -.2ex \@minus -.2ex}%
 {2ex \@plus.2ex}%
{\normalfont\normalsize\protect\baselineskip=12.5pt plus.2pt minus.2pt\bfseries}}
\def\subsubsection{\@startsection{subsubsection}{3}{\z@}%
 {-2.2ex\@plus -.21ex \@minus -.2ex}%
 {1.4ex \@plus.2ex}
{\normalfont\normalsize\protect\baselineskip=12pt plus.2pt minus.2pt\sl}}

\begin{document}
\begin{CJK*}{GBK}{song}
\thispagestyle{empty}
\vspace*{-13mm}
\vspace*{2mm}

\title{3D Multiple Object Tracking on Autonomous Driving: A Literature Review}
\author{ Peng Zhang, Xin Li, Liang He, Xin Lin$^\dagger$,
\\ School of Computer Science and Technology, East China Normal University } 
\renewcommand{\thefootnote}{\fnsymbol{footnote}} 
\footnotetext[2]{Corresponding author}

\noindent {\small\bf Abstract} \quad  {\small {
3D multi-object tracking (3D MOT) stands as a pivotal domain within autonomous driving, experiencing a surge in scholarly interest and commercial promise over recent years. Despite its paramount significance, 3D MOT confronts a myriad of formidable challenges, encompassing abrupt alterations in object appearances, pervasive occlusion, the presence of diminutive targets, data sparsity, missed detections, and the unpredictable initiation and termination of object motion trajectories. Countless methodologies have emerged to grapple with these issues, yet 3D MOT endures as a formidable problem that warrants further exploration. This paper undertakes a comprehensive examination, assessment, and synthesis of the research landscape in this domain, remaining attuned to the latest developments in 3D MOT while suggesting prospective avenues for future investigation. Our exploration commences with a systematic exposition of key facets of 3D MOT and its associated domains, including problem delineation, classification, methodological approaches, fundamental principles, and empirical investigations. Subsequently, we categorize these methodologies into distinct groups, dissecting each group meticulously with regard to its challenges, underlying rationale, progress, merits, and demerits. Furthermore, we present a concise recapitulation of experimental metrics and offer an overview of prevalent datasets, facilitating a quantitative comparison for a more intuitive assessment. Lastly, our deliberations culminate in a discussion of the prevailing research landscape, highlighting extant challenges and charting possible directions for 3D MOT research. We present a structured and lucid road-map to guide forthcoming endeavors in this field.
}}

\vspace*{3mm}

\noindent{\small\bf Keywords} \quad \textcolor{blue} {\small 3D Multiple Object Tracking, Deep Learning, Kernel Correlation Filter, Computer Vision, Graph Neural Network} 

\vspace*{4mm}

\end{CJK*}
\baselineskip=18pt plus.2pt minus.2pt
\parskip=0pt plus.2pt minus0.2pt
\begin{multicols}{2}
 
\section{Introduction}
Object tracking is a popular topic within computer vision, finding applications in diverse domains like video surveillance, autonomous driving, gesture recognition, pedestrian tracking, and augmented reality, encompassing both 2D and 3D scenes. Within this field, a fundamental distinction lies between Single Object Tracking (SOT) and Multi-Object Tracking (MOT). MOT, also known as Multi-Target Tracking (MTT), is a challenging task in computer vision involving the analysis of video data to identify and track objects from various classes, such as pedestrians, cars, animals, and inanimate objects~\cite{1}. Unlike SOT, MOT primarily grapples with complex appearance or motion models to address a multitude of real-world challenges, including scale variations, out-of-plane rotations, lighting variations, point cloud sparsity, crowded scenes, and small objects.

In contrast to 3D object detection, the primary challenge in 3D MOT is to identify and establish correspondences between targets across frames within a sequence of images. This process is fraught with difficulties, including overlap and occlusion between targets, re-identification of old and new targets, and matching targets across consecutive frames. Although 3D MOT methods are more intricate than traditional 2D MOTs, there exists a wealth of excellent 2D MOT methods~\cite{2,3,4,5,6} that can be leveraged.
 
Over time, 3D MOT approaches have evolved rapidly, catering to various sensors, including Lidar-based, camera-based, and multi-modal systems. However, systematic comparisons between these methods have been lacking. Thus, there is a compelling need for a comprehensive assessment of the strengths and weaknesses of these approaches, along with insightful research directions. To address this gap, we propose an exhaustive review of 3D MOT methods, specifically tailored for autonomous driving applications, offering in-depth analysis and systematic comparisons across various categories of approaches.

In this paper, our primary focus centers on vehicle and pedestrian tracking, with a special emphasis on tracking multiple vehicles. Our objective is to present the latest and most pertinent research findings and experimental results in the realm of 3D MOT. This comprehensive perspective aims to equip readers with a systematic understanding of 3D MOT, fostering more profound and forward-looking research in this field.

\begin{figure*}[]
\centering
\includegraphics[scale=0.45]{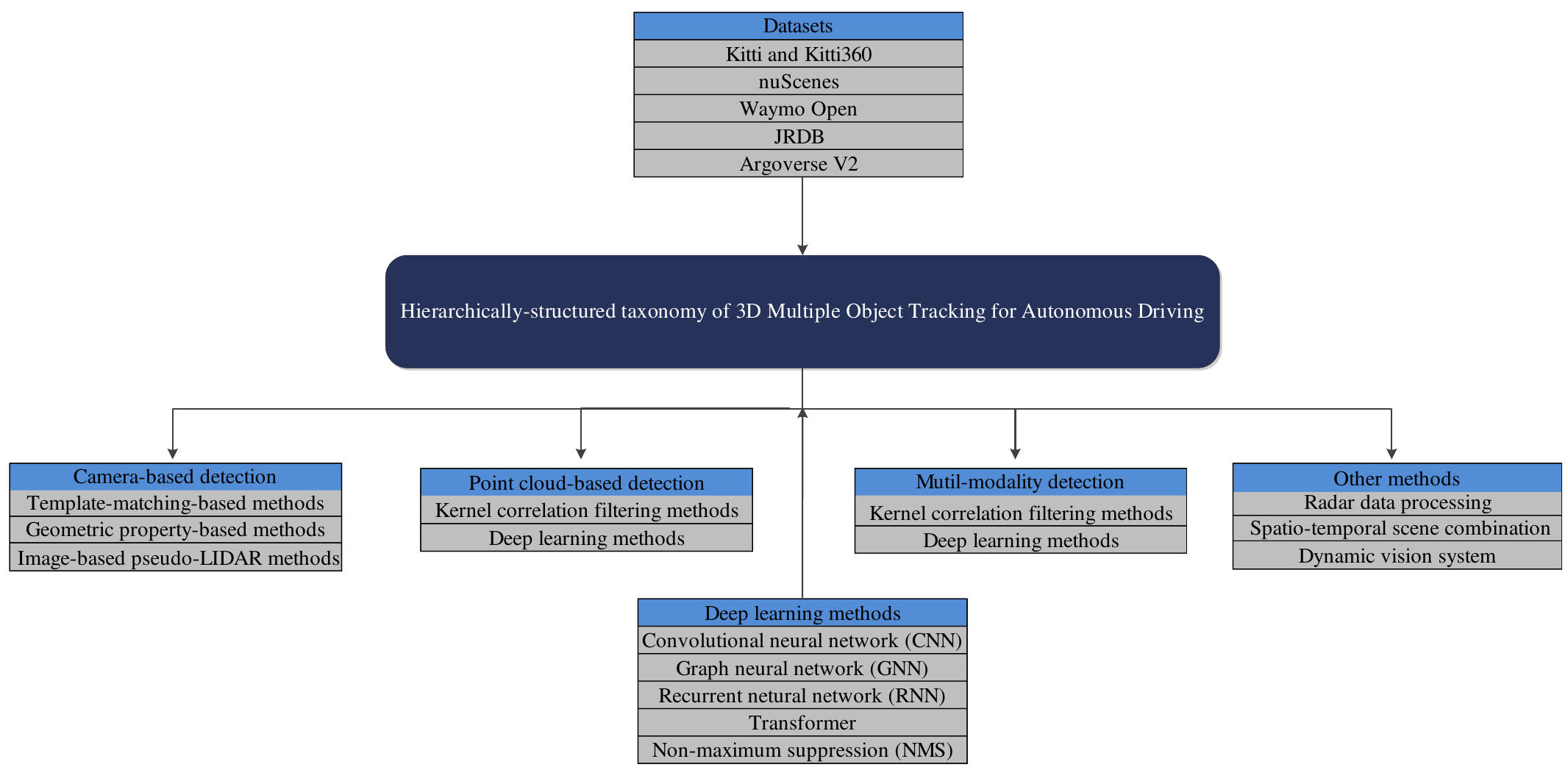}
\caption{\label{Fig1}
3D MOT general structure diagram.}
\end{figure*}

This paper stands as the inaugural comprehensive review of 3D MOT, offering a comprehensive account of its theoretical foundations, experimental aspects, and future developmental trajectories. The paper is organized as follows:

First, we delve into the landscape of related work in 3D MOT in Section~\ref{sec2}, encompassing 3D object detection (Section~\ref{sec2.1}), 3D MOT (Section~\ref{sec2.2}), and the formidable challenges that plague 3D MOT (Section~\ref{sec2.3}). Subsequently, we embark on an in-depth review and analysis of diverse 3D MOT types (Section~\ref{sec3.1}) and approaches (Section~\ref{sec3}), including kernel correlation filtering approaches (Section~\ref{sec3.2}), deep learning methodologies (Section~\ref{sec3.3}), other novel approaches (Section~\ref{sec3.4}), and analysis in 3D MOT methods (Section~\ref{sec3.5}). 

Following this, we delve into the crucial aspects of 3D MOT evaluation, elucidating datasets (Section~\ref{sec4.1}), evaluation metrics (Section~\ref{sec4.2}), and experimental results (Section~\ref{sec4.3}). Lastly, Section~\ref{sec5} provides an overview of the present challenges (Section~\ref{sec5.1}) and offers valuable insights into prospective research directions (Section~\ref{sec5.2}) within the realm of 3D MOT. Figure~\ref{Fig1} offers a clear visual representation of the paper's structure, delineating the primary types, methods, and datasets pertaining to 3D MOT.

\section{Related Work}\label{sec2}

\subsection{3D object detection}\label{sec2.1}
In the realm of 3D MOT, it's crucial to acknowledge that the foundation of current research extensively relies on the outcomes of 3D object detection. In essence, many of the inputs for 3D MOT are derived from the outputs of 3D object detection~\cite{detr3d,ma2023detzero,second,pointrcnn,pvrcnn,pvrcnn++,fan2022embracing}. Therefore, it becomes imperative to explore the landscape of 3D object detection in a comprehensive manner, as it serves as the cornerstone for advancements in 3D MOT.

A prevalent approach in contemporary 3D object detection is the utilization of point clouds or multi-modal data~\cite{multi_survey,huang2022multi,logonet,hmfi,uvtr,bevfusion}. Point clouds represent the real world as a set of discrete points, where each point encapsulates essential data pertaining to an object. In essence, point cloud data embodies an assemblage of vectors in a three-dimensional coordinate system.

It's universally recognized that images inherently possess dense and structured information, encompassing vibrant color palettes and intricate texture details. Nonetheless, images do come with inherent limitations, particularly concerning scale variations induced by objects' varying distances and proximities. In stark contrast, point clouds proffer an abundance of three-dimensional spatial insights, encompassing geometric structures and depth data. Consequently, Lidar point cloud-based methodologies have garnered acclaim for delivering heightened detection accuracy and efficiency when juxtaposed with their camera-based counterparts. A prominent and ongoing area of exploration within 3D object detection and tracking revolves around the integration of multi-modal data to augment detection and tracking precision.

The nexus between point clouds and image fusion pivots on the establishment of absolute coordinates. Essentially, determining the transformation matrix between Lidar and camera sensors suffices to derive the coordinate transformation between the two sensors' respective coordinate systems. This juncture forms a focal point of investigation and discourse within this paper, where we delve into the intricacies of multi-modal fusion.

Presently, the most prominent methodologies for multi-modal fusion in the realm of 3D object detection can be categorized into three primary domains.
\begin{itemize}[noitemsep,nolistsep]
\item \textbf{Early Fusion:} These approaches involves feature fusion before extracting features from the raw sensor data. It has the advantage of maximizing the use of multimodal information. Notable papers in this area include PointPainting~\cite{7}, PIR-CNN~\cite{8}, and Pointaugmenting~\cite{9}, etc.
\item \textbf{Deep Fusion:} These methods extract features from the original data and then fuse the two feature layers. There are two types of fusion methods: Point-based and Voxel-based. The Point-based method has the inherent advantage of having the same index as the image and no spatial variation, while the Voxel-based method can make more efficient use of the perceptual power of convolution. Representative papers include EPNet ~\cite{10}, 3D-CVF~\cite{11}, UVTR~\cite{uvtr}, LoGoNet~\cite{logonet} and BEVFusion~\cite{bevfusion}, etc.
\item \textbf{Late Fusion:} This fusion method first detects the image and the point cloud separately, presents the Proposals, and only after this does the fusion take place. The two are only correlated at the detection level and the multi-modal data does not need to be synchronized or aligned with other modalities; only the final fusion step requires joint alignment and labeling of the data. It has the advantage of low complexity. Notable papers include CLOCs~\cite{12} and Fast-CLOCs~\cite{13}, etc.
\end{itemize}



\subsection{3D multi-object tracking}\label{sec2.2}

The primary distinction between object detection and object tracking is that object detection involves both localization and classification, while object tracking focuses on the trajectory of the target movement in addition to localization based on object detection. The classification referred to in this paper mainly pertains to target types such as cars (vehicle), pedestrians, and bicycles. Undoubtedly, the vehicle category is of utmost importance. Many comprehensive reviews on 3D object detection~\cite{3dodoutlooks2022,3dodsurvey2022,3dodpdall2023} present detailed experimental results specifically for this category.

Object tracking involves establishing the position relationship of the object being tracked in a continuous video sequence and obtaining its complete motion trajectory. During this tracking process, several challenges may arise, including changes in shape and lighting, variations in target scale, target occlusion, target deformation, motion blur, fast target motion, target rotation, target escape from parallax, background clutter, low resolution, and other phenomena.

Autonomous vehicles are equipped with various sensors, including cameras, radar, and Lidar. These sensors collect a large amount of raw data while the vehicle is in motion and send it to the detector module. In this module, the camera obtains the bounding box, the radar obtains the orientation in polar coordinates and Doppler measurements, and the Lidar obtains point cloud data. This data is then fed into the multi-object tracking module.

Most current research on 3D MOT are based on 3D object detection methods, using both point cloud data or multi-modal fusion approaches. The mainstream research methods have focused on kernel correlation filtering methods and deep learning methods. In recent years, there has been a shift towards deep learning-based approaches~\cite{wu2020motionnet,zhang2022motslam,he20233d}, with the main branches being LSTM, CNN, R-CNN, GNN, and Transformer.

\subsection{Challenges of 3D multi-object tracking}\label{sec2.3}

Object tracking involves establishing the position of a target within a continuous video sequence to obtain its complete motion trajectory. This process presents several challenges, including changes in shape and lighting, variations in target scale, target occlusion, target deformation, motion blur, rapid target movement, target rotation, parallax escape of the target, background clutter, and low resolution phenomena. While MOT aims to identify the targets to be tracked in a sequence of images and correspond the targets in different frames one by one. This process also presents numerous challenges. For instance, there can be overlap and occlusion between targets, re-identification of old and new targets, and matching of targets in consecutive frames.

The effectiveness of 3D object tracking systems is largely determined by the accuracy of 3D object detection. The precision of 3D MOT models can be significantly affected by factors such as scale changes, frequent ID switching, rotation, and lighting changes. Despite extensive research on 3D object detection~\cite{li2020deep,ma20223d,zimmer2022survey,lu2022transformers,wang2021attention,wang2023cost}, practical applications still face numerous challenges. Firstly, there is a need for robustness to object occlusion, truncation, and changes in the surrounding dynamic environment. Secondly, most existing approaches rely on object surface texture or structural features, which can easily lead to confusion. Furthermore, there are significant issues with algorithm efficiency while meeting accuracy requirements. Given that 3D object detection forms the necessary foundation for 3D multi-object tracking, achieving a balance between detection efficiency and accuracy is a critical issue that warrants in-depth exploration in this paper.

From the analysis, it's clear that 3D MOT aims to address the challenges of 3D object detection and MOT~\cite{wang2023multi}. The current technology in this field represents a new direction of development, but also faces a series of issues such as frequent target occlusion, background clutter, point cloud sparsity and density, unknown trajectory start and end times, small targets, apparent similarity, inter-target interaction, and low frame rates. On the other hand, current 3D MOT work tends to prioritize accuracy of development over computational cost and system complexity. This indicates that the 3D MOT problem is more complex and requires further study to understand its classification, principles, algorithms, models, experimental results, and trends. This understanding will aid in the investigation of potential solutions.

 


\begin{figure*}[]
\centering
\includegraphics[scale=0.45]{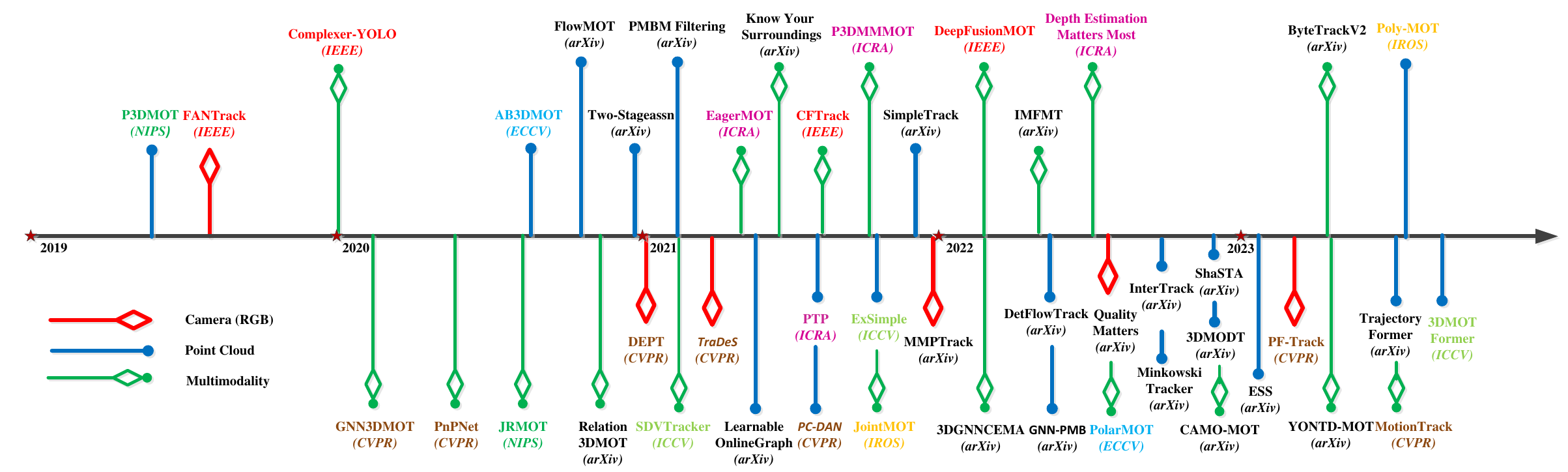}
\caption{\label{Fig2}
A chronological overview of 3D multi-object tracking methods.}
\end{figure*}

\section{3D multi-object tracking approaches}\label{sec3}

\subsection{Types of 3D MOT}\label{sec3.1}

The traditional types of 2D object tracking include Single-Object Tracking (SOT), Video Object Segmentation (VOS), MOT, and Multi-Object Tracking and Segmentation (MOTS). In this paper, we introduce a structured framework for categorizing 3D MOT along two key dimensions. 3D MOT can be approached through various data modalities, including \textbf{camera-based}, \textbf{point cloud-based}, and \textbf{multi-modality-based} methods. Additionally, 3D MOT methods can be broadly classified into two categories: \textbf{Kernel Correlation Filtering methods (KCF)} and \textbf{Deep Learning methods}.

The evolution of object tracking methods has followed a trajectory from classical algorithms to KCF algorithms, and subsequently to deep Learning-based tracking algorithms. Classical tracking algorithms, employed in the early stages, primarily relied on target modeling or tracking target features. This category includes methods such as optical flow and particle filtering.

KCF methods, originally used in signal processing and communication, were adapted for object tracking. These methods found widespread use in real-time tracking systems, contributing to improved tracking efficiency.

With the advent and proliferation of deep learning techniques, entirely new tracking paradigms emerged, employing deep learning methods for object tracking. The choice of robust classifiers plays a pivotal role in achieving high-quality tracking results. Compared to traditional object tracking algorithms like optical flow, the class of algorithms involving correlation filtering tends to offer faster tracking, whereas deep learning-based methods typically deliver superior accuracy. Additionally, factors such as scale adaptation and model update mechanisms significantly influence tracking accuracy.

In our comprehensive analysis, we provide a chronological overview of 3D MOT within the context of autonomous driving (refer to Figure~\ref{Fig2}). It reveals that KCF approaches (\textbf{above the timeline}) and Deep Learning approaches (\textbf{below the timeline}) constitute a substantial portion of the research landscape. In recent years, a prominent trend has been the fusion of point cloud or multi-modal data with deep learning methods. Nevertheless, the KCF method for point cloud data maintains its unique advantages, particularly in terms of speed. It's worth noting that 3D multi-object tracking using Camera-based methods, while having a smaller representation in current literature, predominantly relies on deep learning techniques for its implementation.

\subsection{Correlation filtering and KCF approaches}\label{sec3.2}

The basic idea of Correlation Filter tracking is to design a filter template and use it to correlate with a target candidate region. The position of the maximum output response is considered to be the target position of the current frame. Kernelized Correlation Filter~\cite{14} is an extension of this concept, where kernel functions are introduced in correlation filtering to map the feature space to a higher dimensional space.

In 3D multi-object tracking, the correlation filtering model is divided into four main modules: object detection, motion prediction, association matching, and trajectory management. There are two types of prediction methods commonly used in 3D MOT filtering: the Kalman filter (KF) and the Constant Velocity prediction model (CV). The KF provides smoother predictions when the quality of detection is low, while the CV prediction model is better at handling sudden and unpredictable movements. There are two main approaches commonly used in the association module: IOU-based association and distance-based association. Current matching algorithms take two main forms: one is to formulate the problem as a bipartite graph matching problem using the Hungarian algorithm, and the other is to use a greedy algorithm for matching.

The advantage of the correlation filtering method is that it is faster to perform and the method is relatively simple. However, it relies heavily on the sample information of the last few frames, which accounts for a large portion of the model. If there is inaccurate target localization, occlusion, background disturbance, etc., the fixed learning rate approach will treat these "problematic" samples equally, and the target model will be contaminated and lead to tracking failure.

\subsubsection{Point cloud-based KCF approaches}

AB3DMOT~\cite{21} belongs to the earlier and most important baseline model in the point cloud modal. It is a 3D version of the sort algorithm that uses the principle of minimalism and combines conventional methods to construct a simple, accurate, and real-time 3D MOT system. It has the advantage of being highly accurate and very fast.

The specific process of a 3D MOT baseline model based on a point cloud is as follows:
\begin{itemize}[noitemsep,nolistsep]
\item \textbf{3D object detection:} Input the point cloud data and use the 3D detection module for 3D object detection.
\item \textbf{Use Kalman filtering for prediction:} to estimate the relevant trajectory state (motion trajectory) of the object from the previous frame to the current frame.
\item \textbf{Perform data association matching using the Hungarian algorithm:} the data matching module matches the results of the Kalman filter with the detection results.
\item \textbf{Update track status based on matching results (life-cycle management):} newly appearing objects create tracks, and disappearing objects delete tracks.
\end{itemize}

As can be seen, the 3D MOT algorithm model consists of four modules: detection module, motion prediction module, association module, and trajectory management module. The point cloud modal is categorized by method and can be divided into two groups: kernel correlation filtering methods and deep learning methods. Research on filtering methods using point clouds as an input source is first presented in chronological order.

The most important and typical example of the point cloud filtering method is the aforementioned baseline model AB3DMOT. Probabilistic~\cite{22} addresses some of the problems with the baseline model AB3DMOT by improving it. For example, it replaces the 3D IOU with the Mahalanobis distance as the similarity metric to avoid not being matched because small object tracking and the corresponding detection can easily not overlap. It also uses greedy matching instead of Hungarian matching to achieve better results. Experimental results on the nuScenes validation and test sets show that these methods greatly outperform the AB3DMOT baseline method in terms of the average multi-object tracking accuracy (AMOTA) metric. FlowMOT~\cite{23} combines point motion information with traditional matching algorithms to enhance the robustness of motion prediction. The main process is to use a scene flow estimation network to obtain the implicit motion information between two adjacent frames and compute the predicted detection for each old trajectory in the previous frame. The Hungarian algorithm is then used to generate the best matching relationship, and an ID propagation strategy is used to complete the tracking task. This method can work stably in a variety of situations where filter-based methods may fail. A two-stage data association method~\cite{24} was introduced that had been successful in image-based tracking to a 3D setting, thus providing an alternative to data association for 3D MOT. Two months later, another author proposed the PMBM Filtering method~\cite{25}. This article proposed a PMBM filter to solve the orthogonal MOT problem in autonomous driving applications, which is the first attempt to combine RFS-based methods with 3D Lidar data for MOT applications. The RFS-based tracker outperforms many state-of-the-art deep learning and Kalman filter-based methods. The results of the experiment suggest great potential for further exploration of RFS-based frameworks for 3D MOT applications. The clear decomposition of the multi-object tracking algorithm into four modules was made: a detection pre-processing module, a motion prediction module, an association module, and a trajectory (life-cycle) management module. Some of the now common practices of these four modules were analyzed and improved to summarize them into a unified baseline framework: SimpleTrack~\cite{26}.

One of the most recent articles in this area is Poly-MOT~\cite{27} and it is an efficient 3D MOT method based on the tracking-by-detection framework that enables the tracker to choose the most appropriate tracking criteria for each object category. Specifically, Poly-MOT leverages different motion models for various object categories to accurately characterize distinct types of motion. It also introduces a two-stage data association strategy to ensure that objects can find the optimal similarity metric from three custom metrics for their categories and reduce missing matches. On the nuScenes dataset, this proposed method achieves state-of-the-art performance with 75.4\% AMOTA.

\subsubsection{Multi-modality-based  KCF approaches}

Traditional point cloud data poses numerous challenges for multi-object detection and tracking, as illustrated by the following examples.
\begin{itemize}[noitemsep,nolistsep]
\item \textbf{Data scarcity:} The sparsity of point clouds makes detection and tracking challenging. Additionally, scanned models are often obscured, leading to potential data loss.
\item \textbf{Sensor noise:} All sensors are subject to noise, which can manifest as point cloud perturbations and outliers.
\item \textbf{Vehicle rotation:} Different point clouds can represent the same car, depending on whether it is turning left or right.
\end{itemize}

In contrast to the multi-modal nature of 3D object detection, the multi-modality in 3D MOT primarily refers to the simultaneous use of point cloud data and images as input sources for object tracking. Point clouds and images, two prevalent types of sensory data in autonomous driving, offer complementary benefits: point clouds enable precise target localization, while images provide rich semantic information. By extracting and fusing features from these two sources, 3D MOT can be performed more efficiently.

\begin{figure*}[]
\centering
\includegraphics[scale=0.65]{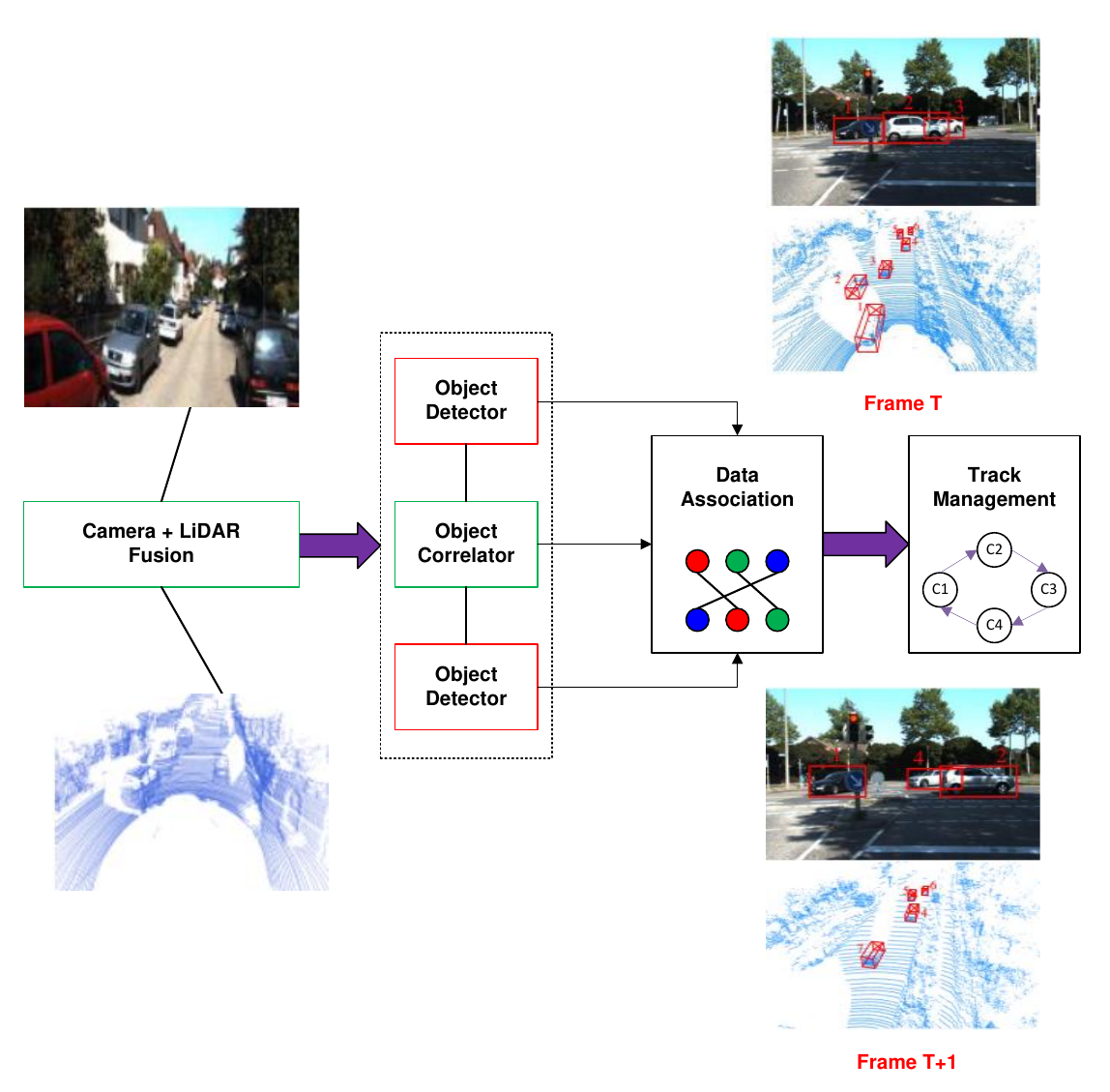}
\caption{\label{Fig3}
An illustration of multi-modal-based 3D MOT fusion approach.}
\end{figure*}

Figure~\ref{Fig3} illustrates the process of 3D MOT using a multi-modal approach. Initially, a continuous image and a point cloud are inputted, and the target is located by their respective detectors. Subsequently, the motion trajectory of the target is predicted, followed by correlation matching. Finally, the trajectory is managed periodically. The image on the right side of the figure graphically displays the tracking results for multiple objects.

Multi-modal methods, such as point cloud, can be categorized into kernel correlation filtering methods and deep learning methods. The research details on multi-modality using filtering methods are presented chronologically.

Complexer-YOLO~\cite{42} is proposed for online semantic segmentation. It represents a novel fusion of state-of-the-art neural network-based 3D detectors and visual semantic segmentation for autonomous driving, marking the first model to combine visual semantics and 3D object detection. Additionally, it introduces Scale-Rotation-Translation scores (SRTs), a fast and highly parametrizable evaluation metric for comparative object detection. This method accelerates inference time by 20\% and reduces training time by half. Most importantly, it applies state-of-the-art online multi-object feature tracking to objective measurements to enhance accuracy and robustness using temporal information. Experiments on KITTI have demonstrated that it maintains a balance between performance and accuracy while operating in real-time.

EagerMOT~\cite{43} is an effective tracking method. It utilizes simple tracking formulas and leverages both the precise point cloud distance measurement and accurate image detection to achieve long-distance tracking. One of its strengths is that it only requires two trained target detectors (2D and 3D), eliminating the need for additional training. Know Your Surroundings~\cite{44} introduced a multi-modal panoramic multi-object tracking framework (MMPAT) that uses 2D panoramic images and 3D point clouds as inputs to infer target trajectories. The method comprises four main modules: a panoramic image detection module, a multi-modal data fusion module, a data association module, and a trajectory inference model. The proposed method was evaluated on the relatively new JRDB dataset and demonstrated satisfactory performance.

CFTrack~\cite{45}, an end-to-end joint object detection and tracking network based on radar and camera sensor fusion. This approach employs a center-based radar and camera fusion algorithm for object detection and a greedy algorithm for object association. The greedy algorithm uses the depth, velocity, and 2D displacement of the detected objects to associate them over time. It makes the tracking algorithm highly effective for obscured and overlapping objects, as the depth and velocity information helps the network distinguish between them. The method is online and has a run-time of 35ms per image, making it highly suitable for autonomous driving applications.

P3DMMMOT~\cite{46} introduces a more advanced multi-modal approach based on P3DMOT. Previous similarity measures for 3D MOT have been Euclidean of centroids or Marxian distances of 3D frames, with the decision to associate or not based solely on distance differences or differences in the size orientation of 3D frames, with little consideration of epistatic features and geometric information. This leads to significantly lower accuracy if the Kalman filter predicts inaccurate position information, for example. Also, the key challenge in improving tracking accuracy is data association and life-cycle management of the tracked target, as these can significantly impact FP and ID Switch metrics. The paper proposes a probabilistic multi-modal, multi-object tracking system with three trainable modules (distance combination, tracking initialization, and feature fusion) that provides data-driven tracking results.

DeepFusionMOT~\cite{47} introduces a robust and fast camera-Lidar fusion-based MOT method that strikes a good balance between accuracy and speed. It designs an effective depth correlation mechanism embedded in the proposed MOT method, leveraging the characteristics of the camera and Lidar sensors. This mechanism allows for tracking objects in the 2D domain and updating the 2D trajectory with 3D information obtained in the Lidar field of view when the object is distant and detected only by the camera, achieving a seamless fusion of 2D and 3D trajectories. IMFMT~\cite{48} uses PointNet++~\cite{49} to obtain a multi-scale depth representation of the point cloud through interactive multi-scale queries and fusion between pixel-level and point-level features. This method enables more distinguishing features to be obtained to enhance the performance of multi-object tracking. Additionally, the paper explores the effectiveness of pre-training and fine-tuning the fusion-based model for every single pattern. Experimental results demonstrate that this approach can achieve good performance on the KITTI benchmark and outperform other methods without using multi-scale feature fusion.

In a related area, recent research includes Depth Estimation Matters Most~\cite{50} and ByteTrackV2~\cite{51}. Depth Estimation Matters Most proposes a multi-level fusion approach that combines different representations (images and pseudo-Lidar) and multi-frame temporal information of objects (tracklets) to improve each object's depth estimation. The paper presents a comprehensive research approach that fuses object detection, SOT, MOT, and various mainstream datasets (Waymo Open dataset, KITTI dataset, and nuScenes dataset) and compares the results. ByteTrackV2 proposes a complementary motion prediction strategy that incorporates detected velocities with a Kalman filter to address abrupt motion and short-term disappearing issues. ByteTrackV2 leads the nuScenes 3D MOT leader board in both camera (56.4\% AMOTA) and Lidar (70.1\% AMOTA) modalities.

\subsection{Deep learning approaches}\label{sec3.3}

In this paper, 3D MOT has more algorithms and models related to deep learning, including LSTM, MLP, CNN, RNN, GNN, and Transformer. Graph clustering methods and GNN focus on aggregating information across frames and objects, while LSTM and RNN focus on improving association performance with motion cues. LSTM and GNN have relatively more applications, and the more promising one at present is the research on moving from 2D and 3D SOT to 3D MOT using transformer.

The 3D MOT deep learning framework, similar to the filtering part, is divided into five main steps:
\begin{itemize}[noitemsep,nolistsep]
\item \textbf{Object Detection stage:} This stage involves analyzing the input frames and using a bounding box to locate the object in the series of frames.
\item \textbf{Motion Prediction stage:} This stage involves analyzing the detection to extract appearance, motion, or interaction features.
\item \textbf{Affinity calculation phase:} The extracted features are used for the similarity distance calculation between detection pairs.
\item \textbf{Association phase:} The similarity/distance metric is utilized in association by providing the same ID to detect corresponding to the same target.
\item \textbf{Life Cycle phase:} This stage manages a series of movement states such as the beginning and end of a goal and its creation and disappearance.
\end{itemize}

Compared with the traditional methods, the 3D MOT tracking algorithm based on deep learning can better adapt to environmental changes, and improve tracking accuracy and tracking speed. The future research direction is to further improve the robustness and accuracy of tracking and can cope with more complex scenes.

\subsubsection{Camera-based deep learning approaches}

Cameras typically use monocular or stereo images as input to predict 3D target instances. There are generally three approaches to this: template-matching-based methods, geometric property-based methods, and image-based pseudo-Lidar methods.

Early tracking methods related to images generally utilized convolutional neural networks (CNNs) for data association in a framework for tracking by detection. Deep learning was used to formulate the data association problem as inference in CNNs, an example of which is FAN-Track~\cite{15}.

Most modern MOT systems follow the paradigm of tracking by detection, which consists of a detector followed by a method of associating the detector to a track. However, this increases complexity and affects speed. As a result, DEFT~\cite{16} is an efficient joint detection and tracking model that relies on the joint learning of an appearance-based object-matching network and an underlying object detection network, with the addition of an LSTM to capture motion constraints. DEFT raises the bar for monocular 3D tracking challenges in the nuScenes dataset, more than doubling the performance of previous top-tier methods. In the same period, a new online joint detection and tracking model, TraDeS (Track to Detect and Segment)~\cite{17}, has been proposed. TraDeS uses tracking cues to assist in detecting end-to-end tracking and infers object tracking offsets from cost quantities, which are used to propagate previous object features to improve current object detection and segmentation. MMPTRACK~\cite{18} provided a large-scale, densely annotated multi-camera tracking dataset in five different environments with the help of an automated annotation system. The system uses overlapping and calibrated depth and RGB cameras to build a high-performance 3D tracker that automatically generates 3D tracking results. This dataset provides a more reliable benchmark for multi-camera, multi-object tracking systems in cluttered and crowded environments. These cues can be unreliable due to visual noise (occlusion and blur) and object properties (position, size, velocity, appearance, etc.), leading to bottlenecks in tracking performance. To address these issues, Quality Matters~\cite{19} effectively guides the network to estimate the quality of predicted object attributes with a quality-aware object association (QOA) strategy. This strategy uses quality scores as an important reference factor for achieving robust associations. Despite its simplicity, extensive experiments show that this strategy significantly improves tracking performance and reduces the performance gap between camera-only and Lidar-based trackers.

The latest article, PF-Track~\cite{20}, proposes an end-to-end multi-camera 3D multi-object tracking (MOT) framework. This framework emphasizes spatio-temporal continuity and integrates both past and future reasoning for tracked objects. On the nuScenes dataset, this method improves AMOTA by a large margin and remarkably reduces ID-Switches by 90\% compared to prior approaches, which is an order of magnitude less.

The camera-based 3D MOT approach has the advantage of low cost and relatively high operational efficiency, but lower accuracy than point cloud and multi-modality approaches. However, it is no longer a mainstream research direction due to the small number of papers.

\subsubsection{Point cloud-based deep learning approaches}

Next, research on deep learning methods using point clouds as input sources is presented in chronological order. Many of these methods use neural learning networks to solve common 3D MOT problems such as high complexity, low accuracy, and missed detection by combining multiple information and relationships and introducing optimization methods for feature aggregation, data association, and operational track management. 

Deep learning methods for point clouds as a single data input source emerged relatively late. Two articles were published in April 2021, one is PTP~\cite{28}. It was based on the baseline model AB3DMOT but used only motion features, not epistemic features. Unlike AB3DMOT, the tracking process incorporates neural learning networks, uses a two-layer LSTM and MLP to acquire motion features, and uses two layers of GNNs for feature aggregation (the feature acquisition mechanism of GNNs). As a result, the interrelated nodes are more similar and better distinguishable features are obtained, allowing for more accurate input to the data association task. Diversity sampling techniques are introduced in the trajectory prediction task, and eventually, matching results are obtained using the Hungarian algorithm. A joint training framework for the prediction and tracking of 3D MOT was proposed to accomplish both tracking and prediction tasks. In the same month, there is another article called Learnable Online Graph Representations~\cite{29}. It designs a graph structure to jointly process detection and tracking states in an online manner, using a fully trainable neural message-passing network for data association. This approach provides a natural way to handle track initialization and false positive detection, while significantly improving track stability, and achieving reasonably good results in the publicly available nuScenes dataset.

There was a rather interesting article called PC-DAN~\cite{30}, with a one-page body of content, but it was well-rounded. It centers on the application of point clouds to the Deep Affinity Network (DAN). Object detection has evolved rapidly over the past two years with the development of deep learning, but while data association is still overwhelmingly hand-crafted to combine apparent features, motion information, spatial relationships, Group relationships, etc., DAN uses deep networks to achieve end-to-end apparent feature extraction and data association. To exploit the inherent power of Lidar data, PC-DAN proposes a point network-based 3D MOT approach. Exploring Simple~\cite{31} was published, which focuses on the biggest drawback of the tracking-by-detection approach: the heuristic matching step usually requires the manual design of matching rules and debugging of relevant parameters. The authors attempt to remove the heuristic matching step from the point cloud 3D object tracking task and propose the SimTrack method to replace the traditional Tracking-by-Detection paradigm for integrated detection and tracking of point cloud 3D targets and explore a simple and accurate end-to-end 3D multi-object tracking paradigm. The method can easily voxelized the point cloud using pillar or voxel methods, then extract features using PointNet~\cite{32} and use 2D or 3D convolution operations in the backbone to obtain bird's eye view features. SimTrack integrates object association, dead object removal, and newborn object detection to reduce the complexity of the tracking system. There were several articles on 3D MOT with point clouds combined with deep learning in 2022. Most previous 3D tracking methods optimize object detection and data association independently. These methods make the network structure complex and limit the improvement of MOT accuracy. DetFlowTrack~\cite{33} proposes a 3D MOT framework based on simultaneous optimization of object detection and scene flow estimation. In this framework, a detection-guided scene flow module is proposed to alleviate the problem of incorrect inter-frame assignment. A box-transform-based method for scene flow realistic computation is proposed for more accurate annotation of scene flows, especially in the case of motion with rotation. Experimental results on the KITTI MOT dataset show that the results are more competitive than state-of-the-art methods in the case of extreme motion and rotation. GNN-PMB~\cite{34} proposed an RFS-based tracker based on the online Lidar MOT task, i.e., using the global nearest neighbor GNN-PMB. This tracker is simple to use but allows very competitive experimental results on the nuScenes dataset.

3D object tracking in dense scenarios where associating existing trajectories with new detections remains challenging, as existing systems tend to ignore critical contextual information. With this in mind, InterTrack~\cite{35} introduces an interactive transformer for 3D MOT to produce discriminative object representations for data association. It extracts state and shape features for each track and detection, and efficiently aggregates global information by attention. Learning regression is then performed on each track/detection feature pair to estimate affinities and a robust two-stage data association and track management approach is used to produce the final track. Significant improvements were obtained through method validation on the nuScenes 3D MOT benchmark, particularly in the category of small target object aggregation. There is also a paper on the Minkowski Tracker~\cite{36}, which aims to jointly address object detection and object tracking. The authors address tracking as the second stage of the object detector R-CNN, predicting the probability of assignment to tracks. First, Minkowski Tracker takes as input a 4D point cloud (3D point cloud video) and generates a Spatio-temporal bird's eye view (BEV) feature map via a 4D sparse convolutional encoder network. Next, TrackAlign aggregates the features of the region of interest (ROI) of the track from the BEV features. Finally, Minkowski Tracker updates the tracking and its Confidence Score based on the probability of matching the detection and tracking predicted from the ROI features.

ShaSTA~\cite{37} uses Lidar sensors to learn shape and spatio-temporal affinities between consecutive frames to better reduce the overall number of false-positive and false-negative trajectories. Quantitative evaluation of the method through ablation experiments and the nuScenes tracking benchmark shows that ShaSTA not only predicts accurate and high-quality motion trajectories but also solves many problems such as missed detections due to cluttered scenes and occlusions. Another article in the same month is 3DMODT~\cite{38}. This model exploits temporal information employing multiple frames to detect objects and track them in a single network, thereby making it a utilitarian formulation for real-world scenarios. This network does not require complex post-processing algorithms and processes raw Lidar frames to directly output tracking results. Experimental evaluations indicate the ability of our model to generalize well across datasets. ESS~\cite{39} presents an efficient semi-automated annotation tool that automatically annotates Lidar sequences with tracking algorithms while offering a fully annotated infrastructure Lidar dataset - FLORIDA (Florida Lidar-based Object Recognition and Intelligent Data Annotation). The advanced annotation tool seamlessly integrates MOT, SOT, and suitable trajectory post-processing techniques. In addition, it provides detailed statistics and object detection evaluation results for the dataset in serving as a benchmark for perception tasks at traffic intersections.

The most recent papers on deep learning methods using point clouds as input sources are TrajectoryFormer~\cite{40} and 3DMOTFormer~\cite{41}. TrajectoryFormer is a novel point-cloud-based 3D MOT framework. It combines long-term object motion feature and short-term object appearance feature to create per-hypothesis feature embedding, which reduces the computational overhead for spatial-temporal encoding. Additionally, we introduce a Global-Local Interaction Module to conduct information interaction among all hypotheses and models their spatial relations, leading to accurate estimation of hypotheses. 3DMOTFormer is a learned geometry-based 3D MOT framework building upon the transformer architecture. It uses an Edge-Augmented Graph Transformer to reason on the track-detection bipartite graph frame-by-frame and conduct data association via edge classification. To reduce the distribution mismatch between training and inference, it proposes a novel online training strategy with an autoregressive and recurrent forward pass as well as sequential batch optimization. Using CenterPoint~\cite{yin2021center} detections, this approach achieves 71.2\% and 68.2\% AMOTA on the nuScenes validation and test split, respectively.

\subsubsection{Multi-modality-based deep learning approaches}

In chronological order, let's delve into the details of the study of deep learning methods with multi-modal input modalities.

Like point cloud combined deep learning methods, multi-modal deep learning methods for 3D MOT have only been gradually developed in the last few years. There are quite a few classic and representative papers published in June 2020. One of them is GNN3DMOT~\cite{52}, a baseline model in the direction of a graph neural network. GNN3DMOT was published in 2020, and it represents a more typical multi-modal and deep learning online 3D MOT framework that can be used as a baseline model for 3D MOT deep learning and graph neural network in particular. It applies GNN to the most landed value 3D multi-object tracking. The algorithm innovates in two ways: firstly, it uses GNN networks instead of the previous feature interaction mechanism to interact the features of the target between multiple targets, making the distinction between different targets greater and reducing the gap between similar targets, thus making the targets more discriminative; secondly, it acquires both 2D and 3D features and fuses them to achieve complementarity of features in different dimensions.

To solve the joint perception and motion prediction problems in the context of autonomous driving, PnPNet~\cite{53} is proposed as an end-to-end model with key components being a multi-object tracker and a novel tracking module. It updates the object trajectory at each time step by solving the data association problem and the trajectory estimation problem to fully capture the temporal characteristics of the target. It is important to note that the entire model is end-to-end trainable and benefits from the joint optimization of all tasks. The authors validated PnPNet on two large-scale autonomous driving datasets (nuScenes and ATG4D), demonstrating better and more accurate prediction capabilities.

JRMOT~\cite{54} is the first application of the new JRDB dataset on 3D MOT. It integrates information from images and 3D point clouds to achieve real-time, state-of-the-art tracking performance. The system is built with neural networks for re-identification, 2D and 3D detection, and tracking description, combined into a joint probabilistic data association framework within a multi-modal recursive Kalman architecture. Another important point is its release of a large-scale 2D+3D JRDB benchmark dataset. Tests have shown that the system can track multiple pedestrians quickly and reliably. To address the problem of tracking without considering feature interactions between objects detected in different frames, Relation3DMOT~\cite{55} first employs a joint feature extractor to fuse 2D and 3D appearance features obtained from 2D images and 3D point clouds, respectively. A new convolution operation, RelationConv, is then proposed to better exploit the correlation between each pair of objects in adjacent frames and to learn a deep affinity matrix for further data correlation. 

SDVTracker~\cite{56} is also a notable paper. Due to the high computational efficiency, many traditional autonomous driving systems use Kalman filters for multi-object tracking, which often rely on hand-designed correlations. However, such methods are less suitable for dense scenarios and multi-sensor modes, often leading to underestimation of state and thus inaccurate predictions. The authors, therefore, propose a practical lightweight tracking system, SDVTracker, which uses deep learning models for correlation and state estimation, combined with interactive multi-model (IMM) filters. The proposed tracking approach is fast, robust, and generalizable across multiple sensor modalities and different VRU classes. Experiments show that the system performs significantly better than hand-designed methods on urban driving datasets, while the CPU runs in less than 2.5ms in scenarios with 100 actors, making it suitable for autonomous driving applications where low latency and high accuracy are critical. 

JointMOT~\cite{57} presents an efficient multi-modal MOT framework with an online joint detection and tracking scheme and robust data correlation for autonomous driving. The work designs an end-to-end deep neural network for joint object detection and association using 2D and 3D data; a powerful affinity calculation module for computing appearance and motion affinities for occlusion perception in 3D space; and a comprehensive data association module for jointly optimizing detection confidence, affinity, and start-end probabilities. Experimental results on the KITTI tracking benchmark show that the proposed method excels in terms of tracking accuracy and processing speed. 3DGNNCEMA~\cite{58} stands out for its exploration of the relatively rare offline 3D MOT. It proposes Batch3DMOT, which follows the paradigm of tracking by detection and represents real-world scenes as directed, acyclic, and category-discontinuous tracking graphs. It proposes a multi-modal graph neural network that uses a cross-edge attention mechanism to mitigate modal intermittency, which translates into sparsity in the graph domain. Attention-weighted convolution on k-NN neighborhoods over frames is also proposed as a suitable means to allow information exchange between disconnected graph components. Extensive experiments have shown that this approach yields an overall improvement of 3.3\% in the AMOTA score of nuScenes and further enhances false positive filtering. 

PolarMOT~\cite{59} encodes 3D detection as nodes in a graph, where spatial and temporal pairing relationships between objects are encoded via local polar coordinates of graph edges. This representation makes the geometric relationships independent of global transformations and smooth trajectory changes, especially under non-global motion. This allows the graph neural network to learn to encode temporal and spatial interactions efficiently and to make full use of contextual and motion cues to obtain the final scene interpretation by associating data with edge classification. In addition to superior performance, PolarMOT has significant generality across locations (Boston, Singapore, Karlsruhe) and datasets (nuScenes and KITTI). CAMO-MOT~\cite{60} is a fairly new study proposing a new multi-modal 3D MOT framework based on appearance-motion optimization CAMO-MOT. which efficiently achieves stable tracking by fusing camera and Lidar information. A motion cost matrix based on confidence scores is also proposed to improve localization and object prediction accuracy in 3D space and to eliminate false detection. The study is the first attempt to introduce multi-category loss to enable multi-object tracking in multi-category scenarios. In addition, it achieves the lowest IDS among all algorithms in the KITTI test set and the top ten algorithms in the nuScenes test set, demonstrating its excellent safety and stability.

There are two relatively new articles: YONTD-MOT~\cite{61} and MotionTrack~\cite{62}. YONTD-MOT is a new MOT framework using a multi-modal fusion method. By integrating object detection and multi-object tracking into the same model, this framework avoids the complex data association process in the classical TBD paradigm and requires no additional training. Confidence of historical trajectory regression is explored, possible states of a trajectory in the current frame (weak object or strong object) are analyzed and a confidence fusion module is designed to guide non-maximum suppression of trajectory and detection for ordered association. Extensive experiments are conducted on the KITTI and Waymo datasets. MotionTrack proposed an end-to-end transformer-based MOT algorithm with multi-modality sensor inputs to track objects with multiple classes. Its objective is to establish a transformer baseline for the MOT in an autonomous driving environment. The proposed algorithm consists of a transformer-based data association (DA) module and a transformer-based query enhancement module to achieve MOT and Multiple Object Detection (MOD) simultaneously. The MotionTrack and its variations achieve better results (AMOTA score at 0.55) on the nuScenes dataset compared with other classical baseline models, such as AB3DMOT, CenterTrack, and probabilistic 3D Kalman filter.
 
\subsection{Other novel approaches}\label{sec3.4}
In the realm of 3D Multi-Object Tracking (MOT), extending beyond the methods introduced above, there are additional techniques that can be harnessed for tracking tasks. These include radar data processing, spatio-temporal scene integration, and dynamic vision systems, among others.

One intriguing approach in the field of 3D MOT involves the fusion of Camera-Radar or Lidar-Radar for target detection. This approach is particularly captivating because the fusion of a camera with low-cost radar can provide precise long-range measurements. Radar TrackNet~\cite{63} is an exemplary study in this domain and it introduced a deep neural network architecture, which employs radar point clouds from multiple time steps to detect road users and compute their tracking information. Nevertheless, it's noteworthy that existing research frequently segregates object detection and object association for enhancement, potentially resulting in inconsistencies in multi-object tracking. An effective method for enhancing tracking accuracy is one grounded in spatio-temporal consistency. For instance, Spatio-Temporal Object Tracking (SpOT)~\cite{64} reformulates tracking as a spatio-temporal issue by representing tracked objects as sequences of time-stamped points and bounding boxes over an extended temporal history. This approach ameliorates the precision of location and motion estimates for tracked objects through learned refinement across the complete sequence of object history. By jointly considering time and space, this representation naturally incorporates fundamental physical principles such as object permanence and temporal consistency. This spatio-temporal tracking framework attains state-of-the-art performance on benchmarks like Waymo and nuScenes. While MOTSLAM~\cite{65} is a dynamic visual Simultaneous Localization and Mapping (SLAM) system configured for monocular tracking of both poses and bounding boxes of dynamic objects. It initially conducts multiple object tracking in association with both 2D and 3D bounding box detection to establish initial 3D object representations. Subsequently, neural network-based monocular depth estimation is applied to infer the depth of dynamic features. Finally, camera poses, object poses, and both static and dynamic map points are jointly optimized using a novel bundle adjustment. The experiments on the KITTI dataset demonstrate that this system has achieved the highest performance in terms of both camera ego-motion and object tracking within monocular dynamic SLAM.

\subsection{Analysis in 3D MOT methods}\label{sec3.5}

In the current phase, detection-based tracking remains the prevailing approach in the field of 3D Multi-Object Tracking (MOT). Substantial enhancements in tracking outcomes have been achieved thanks to recent advancements in 3D object detection algorithms. Nonetheless, the precision of these methods remains limited by the characteristics of the sensors employed. LiDAR is a commonly utilized sensor for data acquisition in 3D MOT; however, the point cloud generated by LiDAR tends to exhibit significant sparsity at extended distances (>80m) due to inherent beam and distance constraints. This inherent sparsity limitation poses a substantial challenge when it comes to accurately detecting and localizing objects.  In contrast, image-based detection can deliver high-quality results even over long distances, thanks to the camera's high pixel resolution. Additionally, current 2D detectors can effectively handle occlusion problems. Therefore, it is crucial to explore 3D MOT methods based on sensor fusion (multi-modal).

Early 3D Multi-Object Tracking (MOT) methods relied on conventional clustering detection techniques, including down-sampling and clustering. However, these methods are susceptible to noise since they rely on the bounding box computation process to obtain detection results. The precision and estimation of bounding box dimensions involve real values with relatively significant errors. With the emergence and growing effectiveness of end-to-end detection methods, the detection component of 3D MOT has gradually transitioned to these techniques.

3D MOT presents various challenges such as target occlusion and target loss over long distances. Another challenge is correlating data between predicted and observed states, i.e., establishing a cost function between them to find a correspondence between observed and predicted state quantities. Most 3D MOT frameworks tend to use the Intersection over Union (IOU) between 3D frames for data correlation, which is indeed a simple and effective method. However, this approach can easily overlook connections between other features, and there may be cases where there are no overlapping regions. Therefore, some data association methods based on Euclidean distance or Mahalanobis distance combined with the Hungarian algorithm or greedy algorithm have been proposed.

From the comprehensive analysis and classification presented above, it's clear that the primary considerations in current 3D MOT algorithms include:

\begin{itemize}[noitemsep,nolistsep]
\item Minimizing or eliminating additional training tasks.
\item Integration of appearance features, motion features, and geometric information.
\item Highlighting the addition and differentiation of background information.
\item Reducing or eliminating manually designed associations.
\item Ensuring the stability and periodicity of the tracking track.
\item Enhancing the security of trackers, such as research on Kalman filter security patches~\cite{66}.
\item Addressing obscure problems~\cite{67}. 
\end{itemize}

Many studies in the field of 3D Multi-Object Tracking (MOT) predominantly employ network frameworks like CenterPoint~\cite{yin2021center}, PointNet~\cite{32}, and PointNet++~\cite{49} for 3D object detection. While earlier approaches often treated object detection and tracking as separate tasks, recent research has shifted its focus towards jointly addressing these two tasks, underlining the significance of synchronization. Currently, deep learning technology is increasingly utilized, particularly in the context of end-to-end Graph Neural Networks (GNNs). However, there is an overarching trend towards the convergence of various methodological models and algorithms, including emerging approaches such as transformers. Within the realm of 3D MOT research, the primary objectives revolve around tracking accuracy and operational efficiency, with all research endeavors centered on these two fundamental aspects.

\section{3D MOT experimental evaluation}\label{sec4}
In this section, we conduct a systematic comparison and analysis of the 3D MOT approaches for autonomous driving. In Section~\ref{sec4.1}, we introduce and compare some commonly used data sets. In Section~\ref{sec4.2}, We list some common metrics and formulas related to 3D MOTs. In Section~\ref{sec4.3}, we conduct a comprehensive analysis of the tracking performances and the inference speeds of various 3D MOT methods, i.e. LiDAR-based, camera-based, multi-modal approaches, on multiple datasets.

\subsection{Datasets}\label{sec4.1}

A multitude of driving datasets have been compiled to provide multi-modal sensory data and 3D annotations for 3D object detection and 3D MOT. The most prominent 3D MOT datasets related to autonomous driving currently include KITTI~\cite{68}, nuScenes~\cite{69}, Waymo Open Dataset~\cite{70}, JRDB~\cite{71}, Argoverse V2~\cite{72}, etc. 

\textbf{KITTI:} This dataset offers a comprehensive collection of images and point cloud data in BIN format, along with various tools for calibration and annotation. The most significant feature is the txt-terminated annotation data stored in the label's directory. Serving as a fundamental resource for detection and tracking tasks related to autonomous driving, its latest version, KITTI 360, provides an extensive dataset that includes abundant sensory information and complete annotations. It annotates both static and dynamic elements of 3D scenes using rough bounding primitives, effectively translating this information into the image domain. As a result, it provides dense semantic \& instance annotations for both 3D point clouds and 2D images.

\begin{figure*}[]
\centering
\includegraphics[scale=0.8]{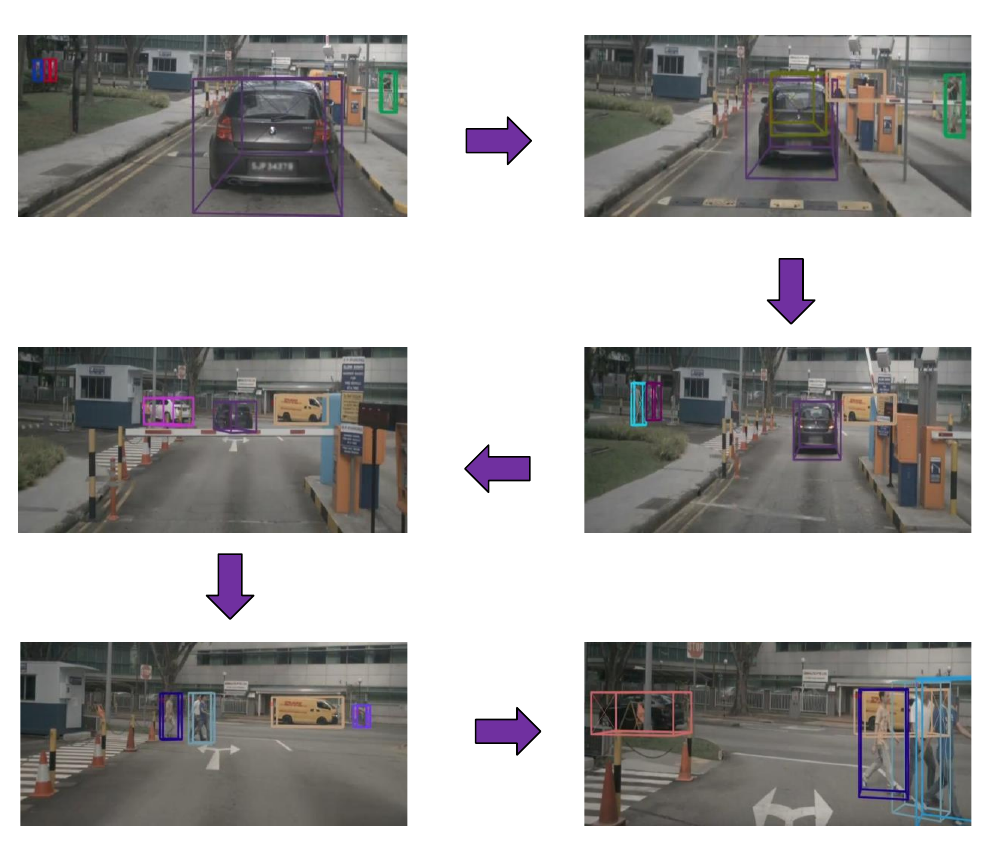}
\caption{\label{Fig4}
The effects of 3D MOT from multiple cameras are demonstrated in the nuScenes dataset.}
\end{figure*}

\textbf{nuScenes:} In addition to a substantial amount of image and point cloud data in PCD format, detailed annotation information is stored across numerous interlinked JSON files. nuScenes has gained widespread acceptance in conference papers and journal publications, establishing itself alongside KITTI as one of the standard mainstream datasets within object detection and tracking research domains. Figure~\ref{Fig4} illustrates a typical application scenario utilizing the nuScenes dataset.

\begin{table*}[]
\centering
\scalebox{0.85}{
\begin{tabular}{ll}
\Xhline{1.2pt}
\multicolumn{2}{c} {\textbf{3D MOT Evaluation Metrics}} \\ \hline
       \textbf{Precision (TP/TP+FP)} & The percentage of correctly identified objects out of the total number of objects identified.
    \\ 
       \textbf{Recall (TP/TP+FN)} & The ratio of correctly matched detection targets to the number of targets given by ground truth.
    \\ 
       \textbf{AP/AR} & Average Precision/Average Recall.
    \\ 
       \textbf{MOTA} & Multi-Object Tracking Accuracy.
    \\ 
       \textbf{MOTP} & Multi-Object Tracking Precision.
    \\ 
       \textbf{IDS} & The number of times a tracking track changes the target label, indicating the number of mismatches.
    \\ 
       \textbf{AMOTA} & Average accuracy: calculated by integrating MOTA for all recall values at different thresholds.
    \\ 
       \textbf{AMOTP} & Average Precision: calculated by integrating MOTP for all recall values at different thresholds.
    \\ 
       \textbf{sAMOTA} & scaled Average MOTA: adjust the AMOTA value in the range between 0\% and 100\%.
    \\ 
       \textbf{FPS} & Frames per second, where a higher value indicates more efficiency.
    \\ 
       \textbf{IDF1} & ID F1 Score. The ratio of correctly identified detections over the average number of GT and computed detections.
    \\ 
       \textbf{FM} & The number of times the track is interrupted.
    \\ 
       \textbf{PT} & Percentage of trajectories tracked by the target segment (1 - MT - ML).
    \\ 
       \textbf{FP/FN} & The total number of false positives(non-existent targets) / The total number of false negatives(missed targets).
    \\ 
       \textbf{MT} & Mostly tracked targets (greater than eighty percent).
    \\     
      \textbf{ML} & Mostly lost targets (less than twenty percent).
    \\ 
       \textbf{Frag} & The total number of times a trajectory is fragmented(i.e. interrupted during tracking).
    \\ 
       \textbf{HOTA} & Higher-order tracking accuracy that balances the effects of performing accurate detection.
    \\  \Xhline{1.2pt}
\end{tabular}}
\newline
\caption{Key Metrics for 3D Object Tracking.}\label{tbl1}
\end{table*}
 
\textbf{Waymo Open Dataset:} Waymo is a relatively recent addition to the available datasets. Despite its novelty, it has already been extensively used for research and evaluation purposes in several new studies.

\textbf{JRDB:} The JRDB dataset is an innovative large-scale 2D and 3D dataset that has recently gained attention due to its utilization in the development, training, and evaluation processes by JRMOT researchers.

\textbf{Argoverse2 (AV2):} Primarily focused on the perception and prediction aspects within the context of autonomous driving systems, AV2 comprises 1000 annotated multi-modal data sequences. These sequences feature high-resolution images captured by seven surround-view cameras, along with imagery output from two stereo cameras. The dataset is the largest Lidar sensor dataset ever created and supports self-supervised learning and emerging point cloud prediction tasks.

\subsection{Evaluation metrics}\label{sec4.2}

In traditional 2D MOT, two key metrics are used to evaluate performance: the Multiple Object Tracking Accuracy (MOTA) and the Multiple Object Tracking Precision (MOTP). MOTA measures the accuracy of the tracking system, while MOTP assesses the precision, which is the closeness of the obtained value to the true value. High scores in both MOTA and MOTP indicate superior performance of the tracking system. Therefore, an ideal tracker would aim to achieve high values in both these metrics.

\begin{equation}
MOTA=1-\frac{\sum_{t}\left(\mathrm{FN}_{t}+\mathrm{FP}_{t}+\mathrm{IDSW}_{t}\right)}{\sum_{t} \mathrm{GT}_{t}}
\end{equation}

The MOTA is a performance metric that evaluates the ability of a system to detect objects and maintain their trajectories. It takes into account three types of errors: False Negatives (FN), False Positives (FP), and Identity Switches (ID Switch). The denominator, GT, represents the number of ground truth objects. An interesting aspect of the MOTA formula is that it calculates a weighted average across all frames, rather than averaging the results of each frame separately. This means that the performance of the detector on each individual frame contributes to the overall MOTA score. MOTA is primarily concerned with the performance of the detector. If the detector has a low rate of False Negatives and Identity Switches, and a high rate of True Positives, then the MOTA score will be higher. This underscores the importance of accurate detection in achieving a high MOTA score.

\begin{equation}
MOTP=\frac{\sum_{t, i} d_{t, i}}{\sum_{t} c_{t}}
\end{equation}

The MOTP is a metric that measures the localization accuracy, which is primarily a reflection of the detector's performance rather than the tracker's. The denominator in the MOTP calculation is the number of successful matches for the corresponding frame. The numerator measures the d-dimensional distance between the detected target and the corresponding ground truth. This distance could be a euclidean distance (where smaller is better) or an overlap rate (where larger is better).

In 3D multi-object tracking, starting from the baseline model AB3DMOT, three important integrated metrics are introduced: Averaged Multi-Object Tracking Accuracy (AMOTA), Averaged Multi-Object Tracking Precision (AMOTP), and sAMOTA. sAMOTA is a scaled accuracy metric, representing a scaled average MOTA. The purpose of sAMOTA is to scale the value of the integrated metric AMOTA to range from 0\% to 100\%. This allows for a more intuitive understanding of the tracker's performance.

\begin{equation}
AMOTA=\frac{1}{n-1} \sum_{r \in\left\{\frac{1}{n-1}, \frac{2}{n-1} \ldots 1\right\}} \operatorname{MOTA}^{\prime}
\end{equation}

\begin{equation}
AMOTP=\frac{1}{n-1} \sum_{r \in\left\{\frac{1}{n-1}, \frac{2}{n-1} \ldots 1\right\}}\frac{\sum_{t, i} d_{t, i}}{\sum_{t} c_{t}}
\end{equation}


 
A comprehensive compilation of evaluation metrics related to 3D Multi-Object Tracking (MOT) is still in progress. After reviewing numerous articles and collecting metrics from various experimental contents, we have assembled the most recent set of 3D MOT evaluation metrics (refer to Table~\ref{tbl1})

Among these evaluation metrics, two key measures are particularly noteworthy: False Positives (FP) and False Negatives (FN). FP refers to points that are incorrectly predicted, while FN, also known as Miss, refers to ground truth points that are not matched.

In addition to these measures, Frames per Second (FPS) and ID-Switches (IDS) are two other important metrics used for evaluating MOT. FPS indicates the rate at which video frames are tracked; a higher rate signifies greater efficiency. This is especially noticeable in point cloud approaches where baseline models using filtering can achieve rates of over 200. IDS is another crucial metric for evaluating MOT, reflecting the number of times that predicted IDs do not match real IDs; lower numbers indicate better performance with zero being optimal. Our analysis suggests that mismatches may occur due to incorrect association or early termination. These insights underline the complexity and challenges inherent in MOT.

\setcounter{figure}{4}
\begin{figure*}[!htb]
\centering
  \subfigure[]{
    \includegraphics[width=8cm,height=6cm]{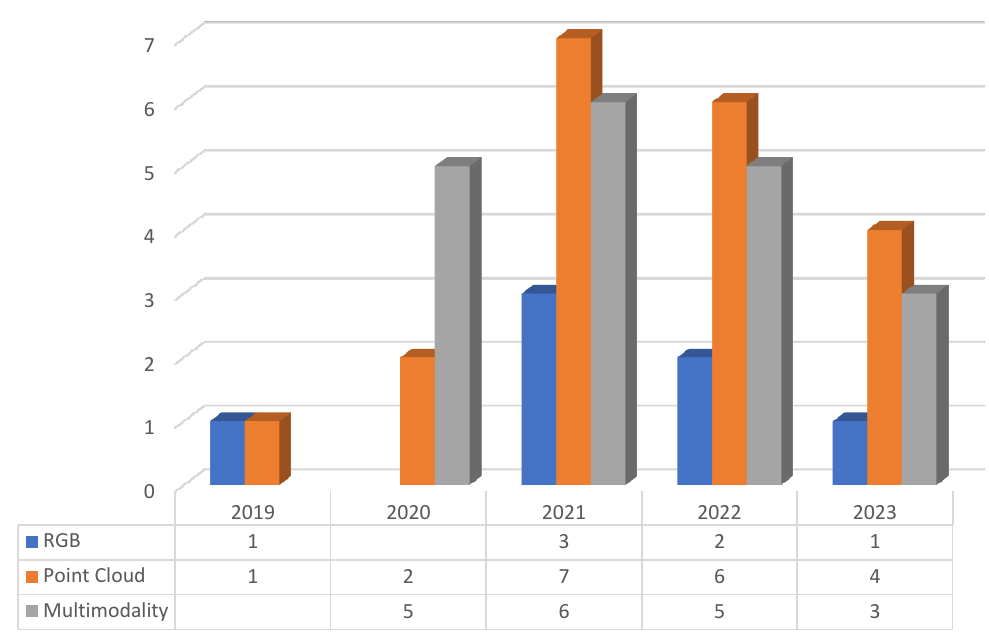}}
  \subfigure[]{
    \includegraphics[width=8cm,height=6cm]{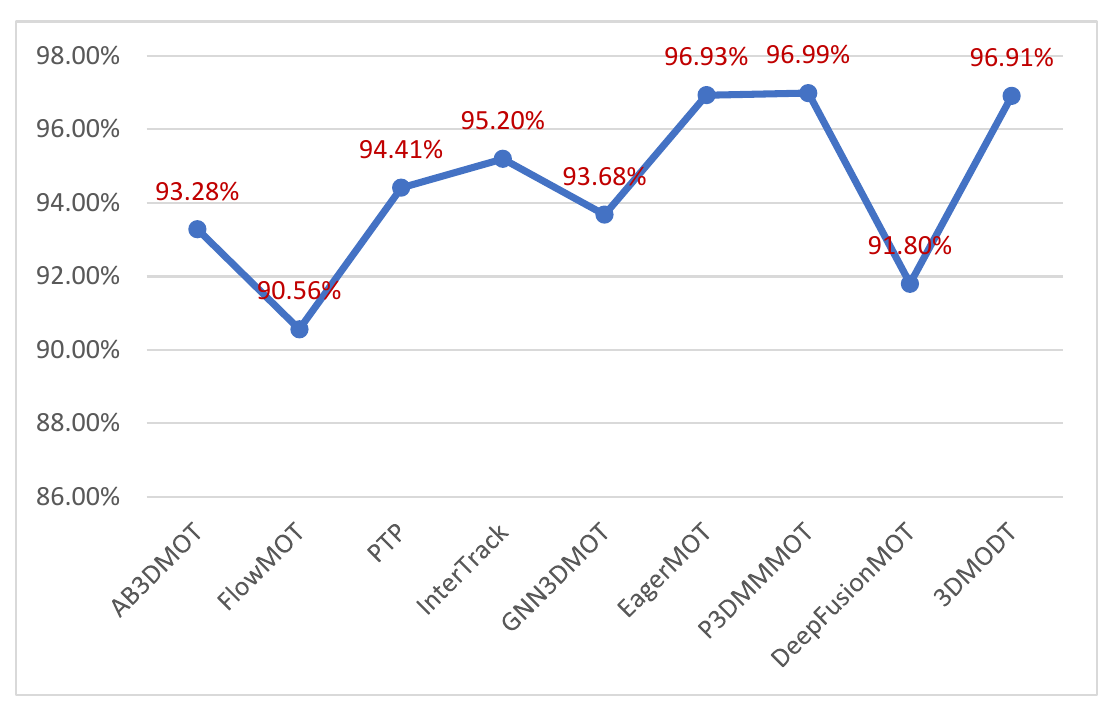}}
  \caption{\label{Fig5}(a) The classification of 3D MOT research results. (b) sAMOTA values for the baseline models and methods.}
\end{figure*}
\baselineskip=18pt plus.2pt minus.2pt
\parskip=0pt plus.2pt minus0.2pt

\subsection{Experimental results}\label{sec4.3}
 
After extensive screening and analysis, we have compiled experimental measurement tables for the categories of Camera, Point Clouds, and Multi-modality. These are presented in Table~\ref{tbl2}. Please pay special attention to the parts of the table highlighted in bold, as these are key areas that need to be summarized.

\begin{table*}[]
\centering
\scalebox{0.68}
{
\begin{tabular}{ccccccccccc}
   \textbf{Results of images method} &&&&&&&&&& \\ \hline
       \textbf{Method} & \textbf{MOTA}       &  \textbf{sAMOTA}      &    \textbf{AMOTA}    &   
       \textbf{IDS}     &   \textbf{FRAG}    &   \textbf{FPS}     &  \textbf{Dataset}      &   \textbf{Category}     &  \textbf{Publication} &   \textbf{Year}    \\ 
    \midrule
       FANTrack~\cite{15} & 0.7772 & - & - &   150  &   872   & - &  KITTI &   Car    &  IEEE &   2019    \\ 
       DEFT~\cite{16} & 0.8810 & - & - &   343     & - & - &  KITTI      &   Car     &  CVPR &   2021    \\ 
       TraDeS~\cite{17} & - & - &    0.118    &   699     & - & - &  nuScenes  &   All     &  CVPR &   2021 \\
       MMPTRACK~\cite{18} & 0.883   & - & - &   1049  & - & - &  MMPTRACK &   Other     &  arXiv &   2021  \\
      PF-Track~\cite{20} & - & - & 0.479 &   181  & - & - &  nuScenes &  All &  CVPR &   2023    \\ \hline
   \textbf{Results of point cloud and KCF method} &&&&&&&&&& \\ \hline
       \textbf{Method} & \textbf{MOTA}       &  \textbf{sAMOTA}      &    \textbf{AMOTA}    &   \textbf{IDS}     &   \textbf{FRAG}    &   \textbf{FPS}     &  \textbf{Dataset}      &   \textbf{Category}     &  \textbf{Publication} &   \textbf{Year}    \\ 
   \midrule
   Probabilistic~\cite{22} & - & - &  0.735 & - & - & - &  nuScenes &   Car   &  NeurIPS  &   2019    \\ 
   AB3DMOT~\cite{21} & 0.8624 & \textbf{0.9328}  &  0.4543  & \textbf{0} &  \textbf{15}  &  \textbf{207}  &  KITTI   &   Car   &  ECCV &   2020   \\ 
   FlowMOT~\cite{23} & 0.8513 & \textbf{0.9056}  &  0.4351 & \textbf{1} & 26 & - &  nuScenes  &  Car &  arXiv &   2020   \\
   Two-Stageassn~\cite{24} & - & - &  0.583  & 512 & 511 & - &  nuScenes   &   All   &  arXiv &   2021   \\ 
   PMBMFiltering~\cite{25} & 0.7167  & - & - &   362   &  1217  & - &  Argoverse  & Car &  arXiv &   2021   \\
   SimpleTrack(10HZ)~\cite{26} & 0.566  & - & 0.668 & 575  & - & - &  nuScenes &   All   &  arXiv &   2021    \\ \hline

    \textbf{Results of multimodality and KCF method} &&&&&&&&&& \\ \hline
       \textbf{Method} & \textbf{MOTA}       &  \textbf{sAMOTA}      &    \textbf{AMOTA}    &   \textbf{IDS}     &   \textbf{FRAG}    &   \textbf{FPS}     &  \textbf{Dataset}      &   \textbf{Category}     &  \textbf{Publication} &   \textbf{Year}    \\ 
   \midrule
   Complexer-YOLO~\cite{42} & 0.757 & - & - & - & - &  100 &  KITTI &   All   &  IEEE  &   2020    \\ 
   EagerMOT~\cite{43} & \textbf{0.9529}  & \textbf{0.9693}  & - & \textbf{1} & - & - &  KITTI & Car &  ICRA &   2021  \\ 
   KnowYourSurroundings~\cite{44} & 0.317 & - & - & 5742 & - & - &  JRDB  &  All &  arXiv &   2021   \\
   CFTrack~\cite{45} & 0.151 & - &  0.2  & - & - & - &  nuScenes   &   All   &  IEEE &   2021    \\ 
   P3DMMMOT~\cite{46} &  \textbf{0.9389}  & \textbf{0.9699} & - & - & - & - &  KITTI  & All &  ICRA &   2021    \\
   DeepFusionMOT~\cite{47} & \textbf{0.913}  & \textbf{0.918}  &  0.4462  & \textbf{1} & - & - &  KITTI  &  All  &  IEEE &   2022    \\ 
   IMFMT~\cite{48} & 0.8424  & - & - &   415   & 568  & - &  KITTI  & All &  arXiv &   2022    \\
   DepthEstimation MM~\cite{50} & 0.4124  & - & 0.1524 & 130  & - & - &  Waymo &  Car  &  ICRA &   2022   \\ \hline
 \textbf{Results of point cloud and deep learning method} &&&&&&&&&& \\ \hline
       \textbf{Method} & \textbf{MOTA}       &  \textbf{sAMOTA}      &    \textbf{AMOTA}    &   \textbf{IDS}     &   \textbf{FRAG}    &   \textbf{FPS}     &  \textbf{Dataset}      &   \textbf{Category}     &  \textbf{Publication} &   \textbf{Year}    \\ 
 \midrule
   LearnableOnlineGraph~\cite{29} & 0.554 & - & 0.656  &   288  & 371 & - &  nuScenes &   All  &  arXiv  &   2021    \\ 
   PTP~\cite{28} &  0.8689  & \textbf{0.9441}  &  0.4615 & \textbf{3} & - & - &  KITTI & All &  ICRA &   2021  \\ 
   PC-DAN~\cite{30} & 0.2256 & - & - & 26009 & - & - &  JRDB  &  All &  CVPR &   2021   \\
   ExSimple~\cite{31} & - & - &  0.836  &  214 & 186  & - &  nuScenes   &   Car   &  ICCV &   2021    \\ 
   GNN-PMB~\cite{34} & - & - & 0.707  & 650    &  345   & - &  nuScenes  & All &  arXiv &   2022    \\
   InterTrack~\cite{35} & - & \textbf{0.952}  &  0.488  & - & - & - &  KITTI  &  Car  &  arXiv &   2022  \\ 
   Minkowski Tracker~\cite{36} & 0.578  & - & 0.698 &   325   & 217  & - &  nuScenes  & All &  arXiv &   2022    \\
   3DMODT~\cite{38} & \textbf{0.9233}  & \textbf{0.9691}   & 0.49  & - & - & - &  KITTI  & Car &  arXiv &   2022    \\
   ShaSTA~\cite{37} & -  & - & \textbf{0.84} & - & - & - &  nuScenes  & Car &  arXiv &   2022    \\ \hline
 \textbf{Results of multimodality and deep learning method} &&&&&&&&&& \\ \hline
       \textbf{Method} & \textbf{MOTA}       &  \textbf{sAMOTA}      &    \textbf{AMOTA}    &   \textbf{IDS}     &   \textbf{FRAG}    &   \textbf{FPS}     &  \textbf{Dataset}      &   \textbf{Category}     &  \textbf{Publication} &   \textbf{Year}    \\ 
 \midrule
   PnPNet~\cite{53} & 0.697 & - & 0.815 &  152   & 310  & - &  nuScenes &   Car   &  CVPR  &   2020   \\ 
   GNN3DMOT~\cite{52} & 0.847  & \textbf{0.9368}  & 0.4527  & \textbf{0} & \textbf{10} & - &  KITTI & Car &  CVPR &  2020  \\ 
   JRMOT~\cite{54} & 0.857 & - & - & 98 & - & - &  KITTI  &  Car &  IROS &   2020   \\
   Relation3DMOT~\cite{55} & 0.804 & - & - & 113  &  265 & - &  KITTI   &   Car   &  arXiv &   2020    \\ 
   SDVTracker~\cite{56} &  0.6944  & - & - & 33118    & 40008    & - &  ATG4D  & All &  ICCV &   2021   \\
   JointMOT~\cite{57} &  0.8627  & - & - &  45 & 586  & - &  KITTI  &  Car  &  IROS &   2021   \\ 
   3DGNNCEMA~\cite{58} & 0.611  & - & 0.713 &  622  & 385  & - &  nuScenes  & All &  arXiv &   2022    \\
   CAMO-MOT~\cite{60} & - & - & 0.763 & 239  & - & - &  nuScenes &   All  &  arXiv &   2022   \\ 
   YONTD-MOT~\cite{61} & 0.8509 & - & - & 42  & - & - &  KITTI &   Car  &  arXiv &   2023  \\  \hline
\Xhline{1.2pt}
\end{tabular}}
\newline
\caption{3D MOT Multi-category Experimental Results and Comparisons.}\label{tbl2}
\end{table*}
 
3D MOT is a relatively new research field, with the majority of results emerging after the second half of 2020, particularly from 2021 to 2023 (as depicted in Figure~\ref{Fig5}a) based on Table~\ref{tbl2}. In terms of publication venues, arXiv papers account for over two-fifths. The predominant datasets used are KITTI and nuScenes, with KITTI dataset being extensively classified (including lane lines, optical flow, scene flow, depth maps) and nuScenes dataset being more specialized. Waymo stands out due to its close relevance to autonomous driving along with newer and more specialized annotation information; whereas JRDB is the most recent dataset that offers exceptional annotations but not exclusively focused on autonomous driving. 
 
From the data in Table~\ref{tbl2}, it can be concluded that:
\begin{itemize}[noitemsep,nolistsep]
\item It should be noted that formal publications on 3D MOT are still relatively scarce as it is a pioneering research topic developed only in the last three or four years; many recent articles can be found on arXiv.
\item The most commonly utilized datasets in the field of autonomous driving include KITTI and nuScenes, while more recent ones consist of JRDB and Waymo datasets.
\item The current research on 3D MOT primarily focuses on vehicles and pedestrians, with a particular emphasis on the vehicle category.
\item The existing challenges in 3D MOT research encompass occlusion, crowding, small targets, and multi-modality issues.
\item Regarding specific metrics, AB3DMOT, GNN3DMOT, EagerMOT, PTP, 3DMODT, and InterTrack exhibit higher sAMOTA values (as depicted in Figure~\ref{Fig5}b). 
\item sAMOTA, AMOTA, and AMOTP are the latest evaluation metrics for 3D MOT that serve as better indicators of model framework performance.
\item Point clouds combined with filtering methods offer relatively efficient results while multi-modal approaches combined with deep learning methods provide greater accuracy. The latter technique is a novel development direction showing promising potential.
\item Some notable baseline models or methods for 3D MOT include \textbf{AB3DMOT}, \textbf{GNN3DMOT}, \textbf{P3DMMMOT}, \textbf{EagerMOT}, \textbf{InterTrack}, \textbf{DeepFusionMOT}, \textbf{3DMODT}, and \textbf{PTP}.
\end{itemize}

From the above analysis, we can draw further insightful conclusions.

Camera-based 3D MOT approaches have garnered attention due to their affordability and rich contextual information. Popular methods include monocular, pseudo-Lidar, multi-camera, and multi-angle approaches. These primarily utilize online efficient joint detection and tracking models, multi-camera tracking datasets, Convolutional Neural Networks, and other deep learning techniques. They offer the benefits of low cost and a relatively simple model structure. However, since images lack the depth information provided by Lidar, ensuring accurate and reliable position and motion cues is challenging. Beyond basic target detection, 3D MOT research concentrates on similarity calculation and data association. This leads to issues such as low MOTA and AMOTA metrics, severe occlusion, and a relative inability to handle cluttered and crowded environments. Despite these challenges, given their low cost and complexity, camera-based 3D MOT approaches remain a significant research direction.

KCF approaches are notable for their low complexity and relatively simple model structure. Experimental results show that not only Lidar-based but also multi-modal-based approaches have high FPS, indicating faster tracking speed. They also have extremely low IDS, indicating a low number of mismatches. Some baseline models (e.g., AB3DMO, FlowMOT, EagerMO, P3DMMMO, DeepFusionMOT) perform exceptionally well. They typically employ effective filters to combine point-wise motion information with traditional matching algorithms, enhancing motion prediction robustness. On the other hand, multi-modality-based KCF approaches focus on more critical aspects such as interactive feature fusion between multi-scale features, effective association strategies, and efficient management of the tracking queue. The fundamental goal of these approaches is to leverage sensor fusion patterns to enhance autonomous driving's accuracy and reliability.

Many traditional autonomy systems use Kalman Filters for multi-object tracking due to their computational efficiency. However, these methods often rely on hand-engineered association and fail to generalize to crowded scenes and multi-sensor modalities. This often results in poor state estimates leading to inaccurate predictions. Deep learning approaches can address these problems relatively well, although their computational efficiency is not as good as the KFC approach and the model structure is more complex. However, they offer advantages such as high prediction accuracy, stable tracking trajectory, and proficiency in handling occlusion and congestion scenarios. Experimental results reveal that the baseline model GNN3DMOT of multi-modal-based deep learning approaches performs excellently. Many 3D MOT research methods related to GNNs are derived from its model structure and methods. Other models and methods such as InterTrack, PTP, 3DMODT, YONTD-MOT also perform exceptionally well. They primarily focus on end-to-end graph neural network design, labeling of new datasets, heuristic matching, integrated detection and tracking, and tracking track stability improvement. These methods can mitigate the impact of false detections in tracking, reduce unnecessary extra training, enable multi-object tracking in multi-category scenes, and achieve robust tracking results. Recently deep learning methods for 3D MOT have increasingly focused on applications in the direction of GNNs, Transformers, and multi-modal applications.
 
\section{Conclusion and future directions}\label{sec5}

In this paper, we comprehensively review and analyze various aspects of 3D MOT for autonomous driving. We start
from the 3D object detection, 3D MOT, and challenges of 3D multi-object tracking, and then we introduce types of 3D MOT and various categories of sensor-based 3D MOT approaches, including KCF, deep learning, camera-based, Lidar-based, multi-modal-based, and other methods. Finally, We further investigate evaluation metrics, datasets, and experimental results for 3D MOT.

Following our extensive and detailed research on 3D MOT, we will delve into an analysis of the prevailing challenges. Additionally, we will forecast the prospective trajectory of advancements in this field.
\subsection{Existing challenges}\label{sec5.1}

 While current research predominantly focuses on 2D MOT approaches encompassing image localization and tracking across multiple cameras, the progress in 3D MOT remains relatively limited, with many methods being derived from upgraded versions of their 2D counterparts. However, leveraging 3D MOT techniques can offer more accurate position and size estimation as well as effective occlusion handling for advanced computer vision tasks. Nevertheless, it necessitates stricter equipment requirements and scene layouts while also entailing higher costs and energy consumption compared to traditional 2D MOT systems.
 
 Contemporary MOT systems typically adhere to the 'tracking by detection' paradigm wherein a detector is followed by an association method for linking detection to tracks. Although combining motion and appearance features has been extensively explored in tracking to enhance robustness against occlusions and other challenges, this often leads to complex and computationally intensive implementations that require further improvements for effectively handling small targets or crowded scenes. Additionally, integrating both 2D and 3D modalities in the context of MOT poses several challenges related to feature alignment, fusion difficulties between different modalities, algorithmic efficiency limitations, as well as high hardware requirements. Research on multi-modal-based 3D MOT approaches involving camera-radar or radar-lidar combinations still remains relatively scarce.

\subsection{Future directions}\label{sec5.2}

Through the previous analysis and summary, we can conclude that 3D object detection plays a fundamental and central role in 3D multi-object tracking. Multi-modal fusion is crucial for achieving accurate 3D object detection and tracking. Additionally, filtering algorithms or deep learning methods are essential for optimizing models in this context. Therefore, it is important to explore potential future research directions.

Recently, the incorporation of deep learning and multi-modality into the field of 3D MOT has emerged as a novel research direction with promising prospects for future investigations, including the following aspects.

\begin{itemize}[noitemsep,nolistsep]
\item \textbf{Multiple classes object tracking (MCOT):} In practical applications, there is generally a need for simultaneous tracking of multiple categories of targets (e.g. sudden animals, balloons, special objects, etc.), especially in the case of dense scenes.
\item \textbf{Multi-Camera and multi-object tracking (MMOT):} The problem of cross-camera data association is also an important research direction.
\item \textbf{Multi-object tracking in combination with other computer vision tasks:} The combinations that can be considered include but are not limited to object segmentation, pedestrian re-identification, human pose estimation, and motion recognition, etc.
\item \textbf{Framework based on tracking and detection joint:} Future methods based on tracking and detection joint will be more applied to the field of 3D MOT.
\end{itemize}

Regarding other details, including the following aspects.From the perspective of methods, deep learning is considered the primary direction for future 3D MOT development, particularly through the integrated application of models and methods such as LSTM, GNN, R-CNN, and scene flow. Many researchers are also focused on enhancing current schemes by designing novel loss functions, network architectures, transformer attention mechanisms, etc. In terms of dataset application, object tracking data annotation is a time-consuming and labor-intensive task; however, a comprehensive annotated dataset can significantly contribute to 3D MOT research and has led to increased recognition of annotators as highly skilled professionals who command higher remuneration. Another crucial research direction involves striking a balance between FPS and metrics like MOTA, AMOTA, and sAMOTA while considering both efficiency and accuracy. Additionally, exploring more efficient multi-modal combined deep learning methods along with conducting comparative analyses across multiple datasets through feature ablation experiments and refining management strategies for 3D MOT tracking trajectories represent meaningful avenues for further investigation.



{\small
\bibliographystyle{unsrt}
\bibliography{egbib}
}





\label{last-page}
\end{multicols}
\label{last-page}
\end{document}